\documentclass[letterpaper]{article} 
\usepackage{aaai25}  
\usepackage{times}  
\usepackage{helvet}  
\usepackage{courier}  
\usepackage[hyphens]{url}  
\usepackage{graphicx} 
\usepackage{natbib}  
\usepackage{caption} 
\usepackage{algorithm}
\usepackage{algorithmic}
\usepackage{amsmath}
\usepackage{amsfonts}
\usepackage{pifont}
\usepackage{color} 
\usepackage{subfigure}
\usepackage{newfloat}
\usepackage{listings}
\DeclareCaptionStyle{ruled}{labelfont=normalfont,labelsep=colon,strut=off} 
\floatstyle{ruled}
\newfloat{listing}{tb}{lst}{}
\floatname{listing}{Listing}
\title{DSRC: Learning Density-insensitive and Semantic-aware Collaborative Representation against Corruptions }
\author{
Jingyu Zhang\textsuperscript{\rm 1,2}\thanks{Equal contributions.} ,
Yilei Wang\textsuperscript{\rm 1}$^{*}$ ,
Lang Qian\textsuperscript{\rm 1},
Peng Sun\textsuperscript{\rm 3}\thanks{Corresponding authors.}, 
Zengwen Li\textsuperscript{\rm 4}\textsuperscript{†},\\
Sudong Jiang\textsuperscript{\rm 4}  ,
Maolin Liu\textsuperscript{\rm 4} ,
Liang Song\textsuperscript{\rm 1,2}\textsuperscript{†}\\
}

\affiliations{
    \textsuperscript{\rm 1} Academy for Engineering and Technology, Fudan University, Shanghai, China \\
    \textsuperscript{\rm 2}
Innovation Platform for Academicians of Hainan Province, Haikou, Hainan, China \\
    \textsuperscript{\rm 3}Duke Kunshan University, Suzhou, China \\
    \textsuperscript{\rm 4}Chongqing Changan Automobile Co., Ltd. Chongqing, China \\
    \tt\small{\{jingyuzhang22, yileiwang23\}@m.fudan.edu.cn}

}

\usepackage{bibentry}

\begin{document}

\maketitle

\begin{abstract}
As a potential application of Vehicle-to-Everything (V2X) communication, multi-agent collaborative perception has achieved significant success in 3D object detection. While these methods have demonstrated impressive results on standard benchmarks, the robustness of such approaches in the face of complex real-world environments requires additional verification. 
To bridge this gap, we introduce the first comprehensive benchmark designed to evaluate the robustness of collaborative perception methods in the presence of natural corruptions typical of real-world environments.
Furthermore, we propose DSRC, a robustness-enhanced collaborative perception method aiming to learn \textbf{D}ensity-insensitive and \textbf{S}emantic-aware collaborative \textbf{R}epresentation against \textbf{C}orruptions. DSRC consists of two key designs: i) a semantic-guided sparse-to-dense distillation framework, which constructs multi-view dense objects painted by ground truth bounding boxes to effectively learn density-insensitive and semantic-aware collaborative representation; ii) a feature-to-point cloud reconstruction approach to better fuse critical collaborative representation across agents.  
To thoroughly evaluate DSRC,
we conduct extensive experiments on real-world and simulated datasets. The results demonstrate that our method outperforms SOTA collaborative perception methods in both clean and corrupted conditions. Code is available at  \url{https://github.com/Terry9a/DSRC}.
\end{abstract}

\section{Introduction}
Perceiving environment accurately is crucial to ensure the driving safety of autonomous vehicles. With recent advancements in Vehicle-to-Everything (V2X) communication technology and intelligent transportation systems, multi-agent collaborative perception has emerged as a promising solution. By exchanging perceptual information, agents can achieve a comprehensive understanding of their surroundings and overcome limitations inherent in single-agent perception, such as limited perception range and obstructed field of view. Recent researches~\cite{v2vnet,xu2022v2x,Where2comm,10549854} have drawn widespread attention to collaborative perception, particularly in autonomous driving. Notably, LiDAR sensors have emerged as a primary focus of state-of-the-art collaborative perception methods~\cite{CoBEVT,yang2023spatiotemporal,disconet} because of their convenient information fusion advantages.

\begin{figure}[t]
  \centering
   \includegraphics[width=1\linewidth]{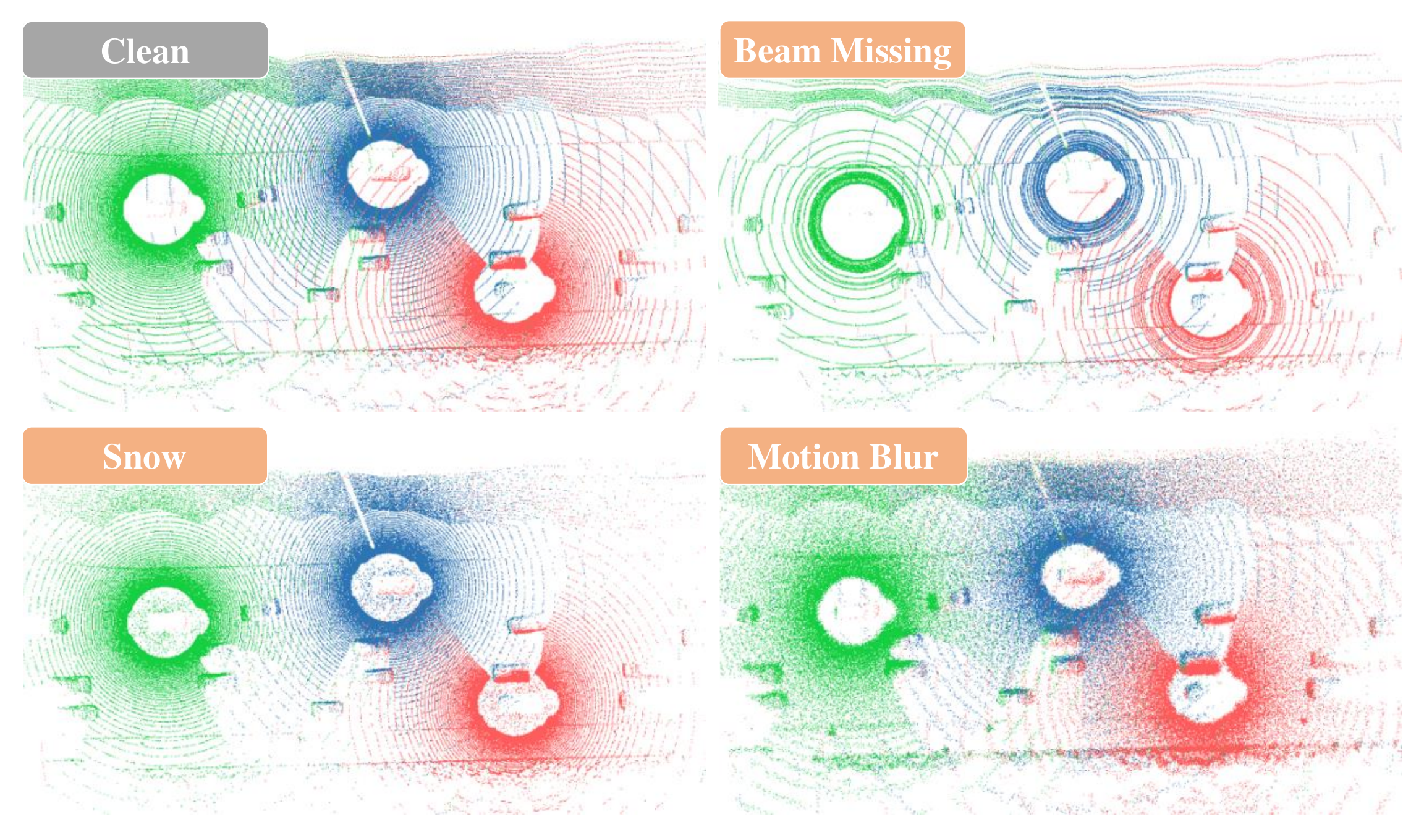}
   \caption{Visualization of typical corruption types in our benchmark. Point cloud from different agents are shown in different colors.
   }
   \label{corruption_types}
   \vspace{-5mm}
\end{figure}

Collaborative perception methods must operate reliably in different geographical locations and different natural environments to ensure the safety of autonomous driving.  While some works~\cite{CoBEVT,wang2023core,wang2023umc,FeaCo} achieve efficient multi-agent feature fusion through well-designed mechanisms to enhance performance within a single domain, the methods often fail in scenarios outside of clean evaluation sets.
Natural corruptions like adverse weather and sensor malfunctions cause issues such as occlusion, signal attenuation, and unpredictable reflections in LiDAR data, leading to loss or misinterpretation of perceptual information. Furthermore, LiDAR's inherent limitations in capturing color and texture details hinder its generalization capability. 
Corruption-induced beam loss or data jitter exacerbates semantic perception difficulties, such as object shape and density. These impairments not only affect each agent's ability to accurately perceive the surrounding environment but also lead to error accumulation in multi-agent perception systems due to the sharing of high-noise perceptual information. Consequently, collaborative perception methods, when exposed to such corrupted scenes, confront risks that are correlated with critical safety concerns.

To address research gaps, we develop a series of common corruption scenarios and establish the first benchmark to comprehensively and rigorously evaluate the corruption robustness of current collaborative 3D object detection methods. Visualization of typical corruption types is shown in Figure~\ref{corruption_types}. These scenarios encompass three distinct corruption sources likely to occur in real-world deployment:
1) Adverse weather conditions, such as fog or snow particles, which obstruct the line of sight of the LiDAR, resulting in sparse object perception and shape degradation~\cite{ren2022benchmarkinganalyzingpointcloud}; 
2) Internal sensor failures, such as crosstalk between multiple sensors, which often creates noisy points within the mid-range areas;
3) External disturbances, such as dust and insects, which cause the LiDAR beam to be missing.
It is essential to proactively address these common corruptions in order to ensure the reliability and robustness of collaborative perception systems.  To achieve this, we propose an innovative distillation framework DSRC to learn \textbf{D}ensity-insensitive and \textbf{S}emantic-aware collaborative \textbf{R}epresentation against \textbf{C}orruptions for robust collaborative perception. It comprises two key aspects:
i) A sparse-to-dense approach is designed to address point sparsity issues in 3D detection under adverse environments. By constructing multi-view dense objects and single-view sparse objects, we establish a sparse-to-dense distillation framework to learn reinforced 3D features in latent space effectively.
ii) A semantic-guided approach is designed to tackle the problem of object semantics degradation in 3D detection under adverse environments. 
Utilizing ground truth bounding boxes to paint teacher point cloud with category semantics, the student model is guided to explore rich semantics effectively, enabling perception of sparse and incomplete objects.
Additionally, we design a feature-to-point cloud reconstruction to regularize feature learning and  better fuse critical collaborative representation across agents.
Our method requires training exclusively on clean data and retains only the student model during inference. Thus, it does not create any additional computational burdens.
To validate the effectiveness of DSRC, we conduct extensive experiments on two collaborative 3D object detection datasets OPV2V~\cite{xu2022v2x} and DAIR-V2X~\cite{dair}. Comprehensive experimental results demonstrate that our method  outperforms previous state-of-the-art methods under any corruption setting.
The main contributions can be summarized as follows:
\begin{itemize}
\item To the best of our knowledge, we conduct the first study on the robustness of multi-agent collaborative perception systems in various corruption scenarios and  establish two corruption robustness benchmarks.
\item We design a sparse-to-dense and semantic-guided distillation framework to enhance the robustness of collaborative perception methods. Additionally, we devise a  point cloud reconstruction module to better fuse critical collaborative representation.
\item We conduct extensive experiments on both real-world and simulated datasets.  The results demonstrate that DSRC outperforms state-of-the-art collaborative perception methods in both clean and corrupted conditions.
\end{itemize}

\section{Related Works}
\subsection{Collaborative Perception}
Collaborative perception enables multiple agents to share complementary perceptual information, promoting a more holistic perception. Based on information transmission and collaboration stages, collaborative perception modes can mainly be organized into early, intermediate, and late fusion. 
Among these, intermediate fusion has garnered increasing attention for its optimal balance between performance and transmission bandwidth. Several intermediate fusion methods for collaborative perception have recently been proposed. 
F-Cooper~\cite{F-cooper} uses an element-wise maximum strategy to aggregate shared features.
V2VNet~\cite{v2vnet} introduces a spatially aware message-passing mechanism for collaborative perception. 
CoBEVT~\cite{CoBEVT} designs a fused axial attention module to capture sparsely local and global spatial interactions across views and agents. CodeFilling~\cite{codefilling} achieves efficient communication by transmitting integer codes instead of high-dimensional feature maps. 
UniV2X~\cite{UniV2X} integrates all key driving modules in a multi-agent system into a unified end-to-end network.
Although these methods show significant performance in multi-agent perception, they are primarily evaluated using standard benchmarks under clean conditions, neglecting common corruptions. This paper addresses this gap by considering the impact of corruptions on multi-agent perception systems.

\begin{figure*}[t]
  \centering
   \includegraphics[width=1\linewidth]{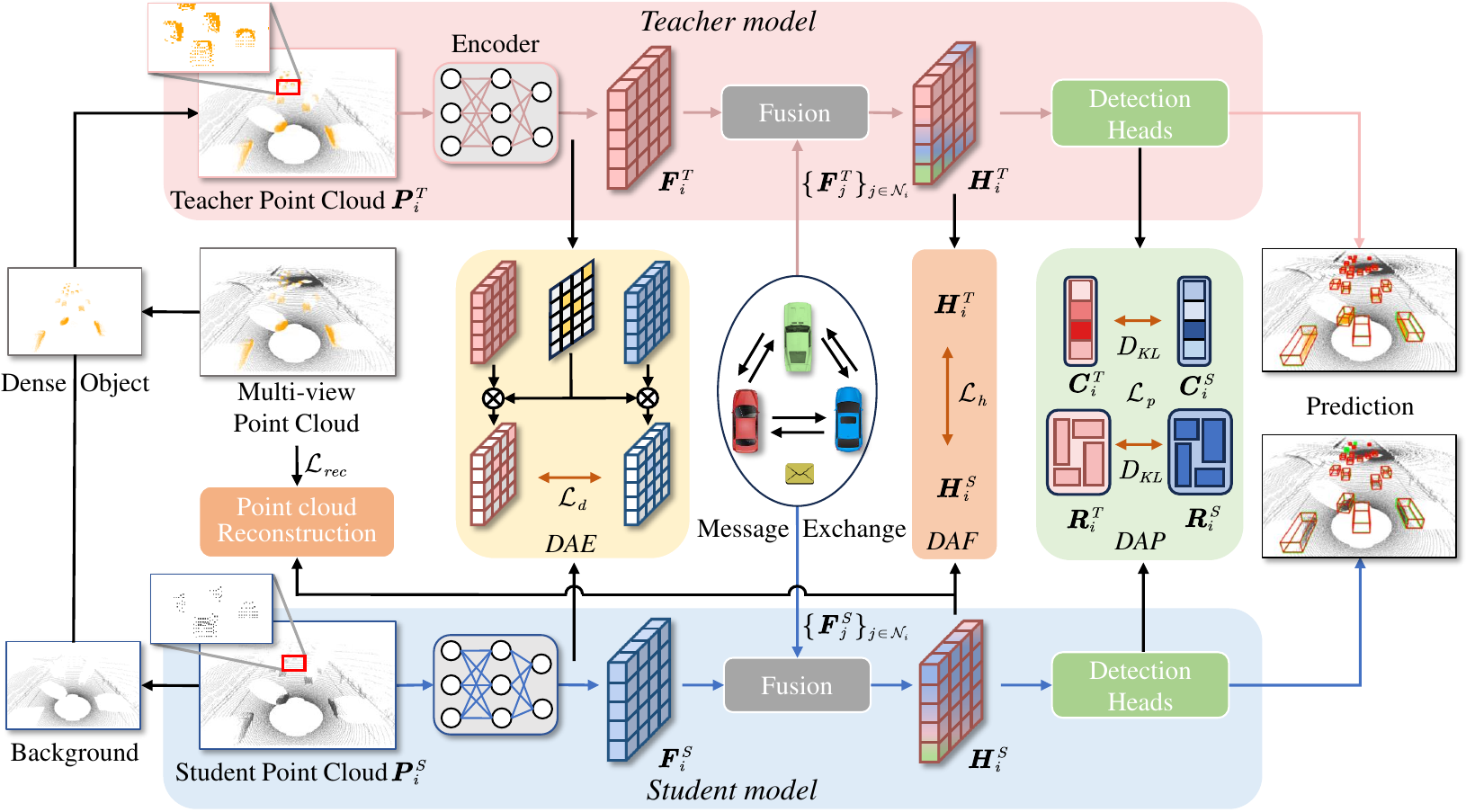}
   \caption{ The overall architecture of the proposed framework DSRC. It contains two branches with identical network structures: Student (Bottom) and Teacher (Top). The framework employs a three-stage distillation strategy: distillation after encoding (DAE), distillation after fusion (DAF), and distillation after prediction (DAP) to achieve effective knowledge transfer and a point cloud reconstruction module to better fuse crucial collaborative representation across agents. During inference, the teacher model and point cloud reconstruction are discarded; only the student model (\textcolor{blue}{blue} data flow) is retained.
   }
   \label{structure}
   \vspace{-3mm}
\end{figure*}

\subsection{Knowledge Distillation}
Knowledge distillation was initially proposed in~\cite{hinton2015distilling} for model compression, aiming to transfer the knowledge learned by a complex teacher model to a simpler student model. Knowledge distillation includes not only label knowledge but also intermediate layer knowledge, parameter knowledge, structured knowledge, and graph representation knowledge, among other forms. Due to its effectiveness, knowledge distillation has been extensively studied in various computer vision tasks, such as object detection~\cite{wang2023object,huang2024knowledge} and semantic segmentation~\cite{wang2024layer,zhu2024saswot}.
BEVDistill~\cite{Bevdistill} projects LiDAR and camera features into the BEV space and adaptively transfers knowledge between heterogeneous representation in a teacher-student fashion.
Ju et al.~\cite{ju2022paint} leverage the semantic information hidden within objects to map semantics onto auxiliary supervisory signals, conveying guiding knowledge to enhance the performance of pure LiDAR models.
Wang et al.~\cite{wang2022sparse2dense} propose a multi-frame to single-frame distillation framework that uses multi-frames to generate dense features as guidance, reinforcing the sparse features of the point cloud in the latent space.
RadarDistill~\cite{bang2024radardistill} considers the sparsity and noisy nature of radar data and develops knowledge distillation to improve the representation of radar data by leveraging LiDAR data.
This paper proposes a three-stage distillation strategy to learn density-insensitive and semantic-aware collaborative representation against common corruptions.

\section{Methods}

In typical driving scenarios involving collaborative perception, corruptions are common. Addressing these corruptions is crucial to ensure the safety of collaborative perception systems. This study aims to achieve robust collaborative 3D object detection through density-insensitive and semantic-aware representation learning. This section introduces the overall architecture, followed by details of each phase.

\subsection{Overall Architecture}
As illustrated in Figure 2, our method employs a teacher-student distillation framework to learn density-insensitive and semantic-aware representation for the student model. 
Considering $M$ agents in the scenario, we first fuse the multi-view point cloud of all agents to obtain a dense  point cloud, representing an overall perspective. Next, we replace object regions in the single-view sparse student point cloud $P^S$ with corresponding multi-view dense object points to obtain the dense teacher point cloud $P^T$. 
We then paint the input point cloud with ground truth labels and use these painted point cloud as input to train the teacher model. Specifically, the point cloud are encoded into Bird's Eye View (BEV) features using a 3D detector with PointPillars~\cite{pointpillars} for feature extraction.  A multi-scale feature attention fusion method~\cite{Where2comm} is then employed to fuse features of all agents.  Lastly, the detection heads predict the classification and regression results. The classification output is the confidence score of the foreground or background, while the regression output is a seven-element tuple $(x, y, z, w, l, h, \theta)$, where $(x,y,z)$ denotes the object center, $(w,l, h)$ defines the 3D box dimension, and $\theta$ represents the heading orientation, respectively.
The student model shares the same structure as the teacher model, including the feature encoder, multi-scale feature attention fusion, and detection head. 
During training, we employ a three-stage distillation strategy to facilitate effective knowledge transfer between the teacher and student models. Additionally, point cloud reconstruction is used to enhance feature extraction and fusion. During inference, only the student model is retained.

\subsection{Teacher Point Cloud Generation}
Due to variations in the locations and viewpoints of multiple agents, significant diversity exists in the observation density and quality within the same spatial region.  For a specific  location in world coordinate, we represent its feature distribution $\mathcal{D}$ as containing all possible features of this specific  location as observed from different angles, with $\boldsymbol{F}_o$ denoting the optimal observable feature. Our objective is to achieve an optimized feature distribution $\mathcal{D}$, defined as follows:
\begin{equation}
\min \sum_{\boldsymbol{F}_j \in \mathcal{D} }\left\| \boldsymbol{F}_j - \boldsymbol{F}_o \right\|^2 .
\end{equation}
In this context, optimal feature is defined as those derived from a multi-view high-density point cloud painted using ground truth bounding boxes. This process ensures that features extracted at low point densities exhibit similarities to those extracted at high point densities, while also guiding towards better semantic information capture and facilitating the convergence of features towards improved representation. Subsequently, we will introduce the multi-view construction of object point cloud and point cloud painting.

\textbf{Multi-view Construction of Object Point Cloud.} 
To obtain the teacher point cloud supervision, the raw point clouds of individual agent are aggregated to generate a comprehensive multi-view point cloud. Then, the object regions in the single-view sparse point cloud are replaced with dense object points derived from multiple views to generate the teacher point cloud.
Specifically, given the raw 3D point cloud $P_i$ collected from the ego agent $i$, and the 3D point cloud of all collaborative agents
$\{\boldsymbol{P}_{j}\}_{j \in \mathcal{N}_i}$ where $\mathcal{N}_i$ denotes the neighbors of the $i$-th agent, we initially transform the 3D point cloud from collaborative agents to the coordinate system of the ego agent $\{\boldsymbol{P}_{j \rightarrow i}\}_{j \in \mathcal{N}_i}=\Gamma_{j \rightarrow i}\{\boldsymbol{P}_j\}_{j \in \mathcal{N}_i}$, where the transformation $ \Gamma_{j \rightarrow i} $ is based on the poses $\xi^{t}_{i}$ and $\xi^{t}_{j}$ of two agents. Then, we aggregate each individual point cloud to construct a multi-view 3D scene: $ \widetilde{\boldsymbol{P}} = \|(\{\boldsymbol{P}_{j \to i}\}_{j \in \mathcal{N}_i}, \boldsymbol{P}_i)$, where $ \|$ represents the aggregate operator. Finally, we replace the object regions in the sparse point cloud $\boldsymbol{P}^S$ with the corresponding dense object points from the multi-view 3D scene to obtain the  dense point cloud $\boldsymbol{P}^T $.

\textbf{Point Cloud Painting}.
Although LiDAR excels in determining the 3D positions of objects, its monochromatic nature limits the acquisition of color and texture information. Additionally, adverse environmental conditions lead to sparse object perception and semantic information degradation, exacerbating the difficulty of object detection. 
To address this challenge, we propose a semantic-guided method to mitigate object semantic degradation in 3D detection under adverse environmental conditions. 
The method guides the student model in extracting richer semantics for the effective perception of sparse and incomplete objects by painting teacher point cloud using ground truth bounding boxes to assign category semantics.
Specifically, considering the original teacher point cloud $ \boldsymbol{P}^T \in \mathbb{R}^{N \times 4}$ , where $ N $ denotes the number of points and the 4-tuple $ (x,y,z,r) $ represents the coordinates and the reflectance intensity. We enhance the teacher point cloud by incorporating a semantic indicator $s$.  Given a point $p_i$, if it lies within the ground truth bounding box, the semantic indicator is set to 1, indicating it as an object point; conversely, the semantic indicator is set to 0, representing a background point. Consequently, we obtain the painted teacher point cloud $ \boldsymbol{P}^T \in \mathbb{R}^{N \times 5}$, where the 5-tuple is $(x,y,z,r,s)$.

\subsection{Density{\footnotesize-}insensitive and Semantic{\footnotesize-}aware Distillation}
We employ a three-stage distillation strategy to achieve effective knowledge transfer: distillation after encoding, distillation after fusion, and distillation after prediction. This strategy forces the student model to match the output representation from the teacher model at various granularities during the encoding, fusion, and prediction processes, thereby reducing the disparities between the two models.

\textbf{Distillation After Encoding.}
Effective feature extraction is pivotal for accurate perception. Thus, we conduct the first stage distillation to align spatial features generated by the teacher and student models, facilitating the student model to produce dense high-quality features. 
Given the $i$-th vehicle local observations $\boldsymbol{P}_i^S$, the extracted features are represented as $\boldsymbol{F}_i^S = \Phi_{\text {enc}}(\boldsymbol{P}_i^S) \in \mathbb{R}^{H \times W \times C}$, where $\Phi_{\text {enc}}(\cdot)$ denotes the feature encoder and $H$, $W$, and $C$ stand for the height, width, and channel of the feature map, respectively. Similarly, the features extracted from the dense teacher point cloud $\boldsymbol{P}_i^T$ are $\boldsymbol{F}_i^T $.
Then, we perform feature constraints by minimizing the $l$2 distance between the two feature maps. To mitigate the effect of background noise, we use labels to create a foreground binary feature mask $\boldsymbol{M}_i\in \mathbb{R}^{H \times W} $, ensuring that the loss computation focuses only on the foreground region. The loss function can be formulated as:
\begin{equation}
\mathcal{L}_d=\sum_{j \in \mathcal{N}_i \cup\{i\}}{\boldsymbol{M}_{j} \cdot\left\|\boldsymbol{F}_{j}^T-\boldsymbol{F}_{j}^S\right\|_2}.
\end{equation}

\textbf{Distillation After Fusion.}
By fusing perception features from all agents, we aim to achieve a highly capable fused representation. High-quality feature fusion is the first step towards holistic perception. Therefore, we perform the second-stage distillation to  align the intermediate fused features $\boldsymbol{H}_i^S$ with $\boldsymbol{H}_i^T$, effectively ensuring consistent integration of each agent's perception throughout the learning process. The distillation loss is formulated as:
\begin{equation}
\mathcal{L}_h=\left \|\boldsymbol{H}_{i}^T-\boldsymbol{H}_{i}^S\right \|_2.
\end{equation}

\textbf{Distillation After Prediction.}
Prediction discrepancies intuitively reflect significant information that distinguishes the student model from the teacher model, and our final goal is to decode the classifications and 3D bounding boxes from the fusion features. Thus, ensuring alignment at the prediction level further contributes to the consistency and accuracy of results. To this end, we employ Kullback-Leibler (KL) divergence loss to compute prediction discrepancies between the teacher and student, aiming to transfer deep knowledge by minimizing prediction differences between them. 
Given the class decoding outputs $\{\boldsymbol{C}_i^T,\boldsymbol{C}_i^S\}$ and regression decoding outputs $\{\boldsymbol{R}_i^T,\boldsymbol{R}_i^S\}$ for the teacher model and the student model. The prediction level distillation can be formulated as:
\begin{equation}
\mathcal{L}_p=D_{KL}\left(\boldsymbol{C}_i^T, \boldsymbol{C}_i^S \right)+  D_{KL}\left(\boldsymbol{R}_i^T, \boldsymbol{R}_i^S \right).
\end{equation}
In summary, the total loss of distillation is formulated as:
\begin{equation}\mathcal{L}_{kd} = \alpha \cdot \mathcal{L}_d + \beta \cdot \mathcal{L}_h + \gamma \cdot \mathcal{L}_p,
\end{equation}
where $\alpha$, $\beta$ and $\gamma$ are balance hyperparameters. 
\begin{figure}[h]
  \centering
   \includegraphics[width=0.99\linewidth]{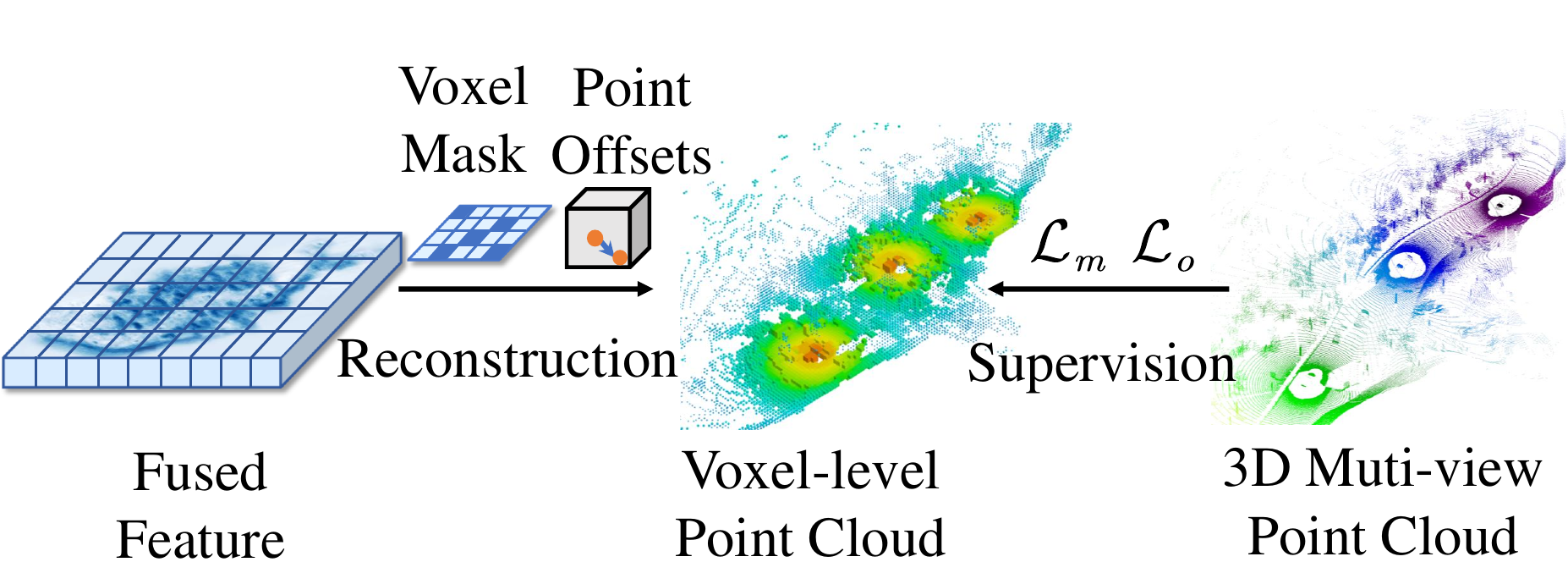}
   \caption{Illustration of the proposed  point cloud reconstruction module. It provides additional supervision to better fuse critical collaborative representation across agents.}
   \label{rec}
\end{figure}

\subsection{Point Cloud Reconstruction }
While distillation between the teacher and student models facilitates knowledge transfer, it achieves suboptimal performance because the model relies solely on task-specific losses to indirectly learn collaborative representation fusion.
Therefore, we supervise the student model on a feature-to-point cloud reconstruction to better fuse critical collaborative representation across agents.

As illustrated in Figure~\ref{rec}, our primary insight is that if the agent successfully acquires a reliable fused feature after message exchange and fusion, the fused feature should be able to reconstruct the complete point cloud scene in reverse. This reconstruction idea is task-agnostic and provides an explicit and sensible supervision of collaboration. However, directly reconstructing large-scale dense point clouds is challenging. Inspired by~\cite{wang2022sparse2dense}, we implement a voxel-level reconstruction strategy, which decouples the point cloud  reconstruction task into two subtasks: occupancy mask  prediction and point offsets prediction for non-empty voxels. The occupancy mask prediction generates a soft voxel occupancy mask $V_m$, representing the probability of a voxel being non-empty. For non-empty voxels, the point offsets prediction  estimates the point offset set $O_p$ for each voxel, representing the offsets from the voxel center $V_c$ to the average input points of the voxel. Therefore, the reconstructed point can be expressed as follows:
\begin{equation}
P_c=\left(O_p+V_c\right) \times V_m.
\end{equation}
The occupancy mask prediction loss $\mathcal{L}_{m}$ and point offsets prediction loss $\mathcal{L}_{o}$ can be expressed as follows:

\begin{equation}
\mathcal{L}_{m} = -\frac{N_b}{N_f} \sum_{j=1}^{H\scalebox{0.6}{$\times$}W} \left( y_j \scalebox{0.9}{$\log$}(p_j) + (1 - y_j) \scalebox{0.9}{$\log$}(1 - p_j) \right),
\end{equation}

\begin{equation}
\mathcal{L}_{o}=\frac{1}{\left|N_f\right|} \sum_i^{N_f}\left|\left(O_{p_i} + V_{m_i}\right)-O_{\text{gt}_i}\right|,
\end{equation}
where $N_b$ and $N_f$ represent the numbers of background and foreground voxels, $p_j$ and $y_j$ stand for the predicted value and ground truth of the voxel mask, $H$ and $W$ stand for the height and width of the mask, and $j$ indexes the voxels in $V_{m}$. The reconstruction loss $\mathcal{L}_{rec}$ is defined as $\mathcal{L}_{m}$  + $\mathcal{L}_{o}$ . During training, the overall loss function is:
\begin{equation}
\mathcal{L}=\mathcal{L}_{\text {detect }}+\mathcal{L}_{kd}+\mathcal{L}_{rec},
\end{equation}
where $\mathcal{L}_{\text {detect }}$ is the detection loss.
\section{Experiment}

\subsection{Datasets and Evaluation Metrics}
\textbf{Datasets.} 
We validate the proposed DSRC in the LiDAR-based 3D object detection task using two main datasets: OPV2V (Xu et al., 2022c) and DAIR-V2X (Yu et al., 2022).
\textbf{OPV2V} is a large vehicle-to-vehicle collaborative perception dataset collected by Carla~\cite{carla}  and OpenCDA~\cite{opencda}. This dataset comprises 11,464 frames of annotated 64-line point cloud and RGB images with 3D annotations. The training/validation/testing splits include 6,764, 1,981, and 2,719 frames.
\textbf{DAIR-V2X} is a real-world dataset for vehicle-to-infrastructure perception. It samples 9K synchronized vehicle and infrastructure LiDAR frames from 100 representative scenes at a frequency of 10Hz. The RSU LiDAR is 300 lines, while the vehicle’s LiDAR is 40 lines. The ratio of training/validation/testing sets is 5:2:3.

\textbf{Evaluation Metrics.}
We adopt the average precision (AP) at the intersection-over-union (IoU) thresholds of 0.5 and 0.7 to evaluate the detection performance. Meanwhile, we also use the mean Corruption Error (mCE) as the primary metric to compare the model robustness following ~\cite{dong2023benchmarking,kong2023robo3d}. The mCE represents the percentage of performance drop as:
\begin{equation}
{CE}_i=\frac{ {AP}_{clean }- {AP}_{i}}{ {AP}_{clean }}, \quad {mCE}=\frac{1}{N} \sum_{i=1}^N {CE}_i,
\end{equation}
 where $AP_{clean}$ denotes the average precision on
the \textit{clean} evaluation set and $N$ is the total number of corruption
types.
\subsection{Experiment Setup}
Due to lacking a suitable robustness evaluation benchmark, existing 3D perception models tend to overfit clean data distributions rather than realistic scenarios. This paper aims to enhance the perception performance of multi-agent systems under unknown corruptions and establish a benchmark for evaluating the robustness of collaborative perception. Assuming a point in a LiDAR point cloud, with coordinates and intensity, we simulate a corrupted point through a mapping, where the mapping rules are constrained by physical principles or engineering experience. We simulate six common types of corruption, including beam missing, motion blur, fog, snow, crosstalk, and cross sensor.  Due to space limitations, a detailed definition and implementation of the corruption simulation algorithm are provided in the \textit{Appendix}.

\begin{table*}[t]
\setlength\tabcolsep{1.8mm}
\renewcommand{\arraystretch}{1.1}
\begin{small}
\begin{tabular}{c|ccccccc}
\hline
Corruptions                         & Clean        & Beam Missing     &Motion Blur      & Fog          & Snow            &Crosstalk     &Cross Sensor \\ \hline
Model/Dataset                             & \multicolumn{7}{c}{OPV2V}           \\ \hline

No Collaboration               & 78.85/65.05  & 63.50/48.01      & 61.99/36.79     &59.88/49.38  & 57.56/45.48      & 73.73/58.47   &60.64/44.00   \\
Late Fusion                         & 87.48/80.41  & 80.40/69.36      & 76.68/51.84     &82.91/69.48  &69.60/58.87       & 82.47/71.98   & 77.43/63.75  \\
F-Cooper~\cite{F-cooper}          & 87.40/79.40    &75.65/65.91   & 73.44/48.30    &63.76/53.52   & 59.60/53.21   & 63.76/53.52 &75.65/65.91  \\
V2VNet~\cite{v2vnet}      &92.29/83.20  & 83.30/70.35    &83.87/65.17   &68.22/57.06 & 73.90/66.93 &90.33/77.68  &81.40/67.53  \\
V2X-ViT~\cite{xu2022v2x}   & 91.49/83.27  & 82.31/70.88  & 80.97/61.18   &70.97/60.97   & 64.96/56.87  & 78.07/66.42  & 80.47/68.02\\
CoAlign~\cite{coalign}      & 91.08/84.61  & 83.00/74.31  & 77.46/58.48      &70.29/64.49  & 66.10/59.95  & 82.28/74.68   &80.87/71.36 \\
ERMVP~\cite{zhang2024ermvp}  & 92.03/85.41	   & 81.85/69.77  & 84.02/67.42      &73.62/65.21 & 68.95/63.63  & 85.85/80.12	  &	79.52/72.78    \\
Mrcnet~\cite{Mrcnet}  & 91.73/83.28   & 83.42/71.88  & 85.71/68.20      &70.17/59.86 & 67.54/62.85  & 81.25/76.47	   &74.29/62.06    \\
CoBEVT~\cite{CoBEVT}  & 91.39/86.18   & 82.07/74.53  & 84.78/67.21      &75.04/68.63  & 68.83/63.00  & 87.35/80.55   & 80.23/73.04    \\
\textbf{DSRC (Ours)} & \textbf{92.58/88.45}   & \textbf{85.82/79.59} & \textbf{86.20/69.41}    &\textbf{83.54/69.84}  & \textbf{74.14/67.25} & \textbf{90.76/84.57}   &\textbf{85.77/77.64}   \\
\hline 
Model/Dataset                           & \multicolumn{7}{c}{DAIR-V2X}           \\ \hline

No Collaboration               & 54.19/42.43 & 43.11/33.62    & 38.74/24.53     &26.19/19.75  & 35.75/26.82      & 47.83/35.05  &30.01/22.70    \\
Late Fusion                    &  56.33/43.47 & 41.35/30.37      & 43.41/23.04     &37.14/25.09  &45.04/28.01 &52.18/33.80      & 33.32/19.95  \\
F-Cooper~\cite{F-cooper}       &58.23/41.74    &40.11/27.76    & 33.71/17.36   &27.03/18.64 & 20.88/11.69    & 50.78/33.97 &26.29/17.48  \\
V2VNet~\cite{v2vnet}         &61.89/44.63  & 46.69/31.05   & 54.68/33.77   &36.69/25.80  &48.25/32.40 & 60.80/42.07 & 31.48/19.67    \\
V2X-ViT~\cite{xu2022v2x}   & 59.70/43.67    & 41.73/29.67   & 44.15/29.53    &31.19/23.32  &37.43/29.05 & 56.29/39.72   & 26.30/17.86    \\
CoAlign~\cite{coalign}      & 68.23/55.08   & 54.31/41.33  & 57.46/41.39      &41.40/32.15  &43.31/31.97   & 61.64/49.54  & 39.31/28.10   \\
ERMVP~\cite{zhang2024ermvp}  & 61.02/46.39	   & 46.59/34.64 & 	49.40/31.80     &35.09/26.94  & 42.36/31.37  & 56.64/41.56  & 28.10/20.00  \\
Mrcnet~\cite{Mrcnet}  & 58.15/43.93   & 44.51/31.82  & 49.67/30.78     &35.81/26.65 & 37.72/24.63	  & 	54.29/39.64  & 30.67/20.82  \\
CoBEVT~\cite{CoBEVT}  & 61.77/45.18  & 46.38/31.93   & 43.16/24.32    &35.39/24.15  & 36.72/24.00  & 58.31/40.31    & 32.35/20.73    \\
\textbf{DSRC (Ours)} &\textbf{69.51/56.54}   &\textbf{55.62/42.62} & \textbf{59.76/42.81}    &\textbf{42.83/33.65}  &\textbf{45.41/33.52} &\textbf{63.99/51.26}   &\textbf{39.81/28.50}   \\
\hline 
\end{tabular}
\end{small}
\caption{Overall performance on  OPV2V and DAIR-V2X datasets under clean and corrupted conditions. The results are reported in AP@0.5/0.7.  
}
\vspace{-4mm}
\label{tab:performance}
\end{table*}

\subsection{Implementation Details}
We implement the proposed and comparative models using the PyTorch framework~\cite{paszke2019pytorch}  and train them on a single RTX 3090 24G GPU using the Adam optimizer~\cite{kingma2014adam}. The cosine annealing learning rate scheduler is used with an initial learning rate of 2e-3. All models are trained for 40 epochs with a batch size of 2 on the original dataset, employing early stopping to identify the optimal epoch. All detection models utilize PointPillars~\cite{pointpillars}  as the backbone to extract 2D features from the point cloud, and  0.4 m width/length is used for each voxel. We assume that all agents have a communication range of 70 m following~\cite{xu2022v2x}. All the agents out of this broadcasting radius of ego agent will not have any collaboration. The balance hyperparameters $\alpha$, $\beta$, and $\gamma$ are 1, 1, and 0.5.  Following existing work~\cite{pointpillars}, we adopt the smooth $L$1 loss for regression and focal loss~\cite{Lin_2017_ICCV} for classification.

\subsection{Quantitative Evaluation}
\textbf{Comparison of Detection Performance.}
Table~\ref{tab:performance} presents the 3D detection performance comparison results based on two datasets. 
We use the No Collaboration method as a baseline, which relies solely on the LiDAR point cloud from the ego agent. Additionally, we compare this with the Late Fusion approach, where detection outputs from all agents are merged, and non-maximum suppression is applied to generate the final results. Furthermore, we evaluate several state-of-the-art methods for the intermediate fusion strategy: FCooper~\cite{F-cooper}, V2VNet~\cite{v2vnet}, V2X-ViT~\cite{xu2022v2x}, CoAlign~\cite{coalign}, and CoBEVT~\cite{CoBEVT}.
We see that DSRC:
i) under clean conditions, all methods achieve acceptable performance, with our method outperforming state-of-the-art collaborative perception methods on both datasets, thus showcasing the superiority of the DSRC collaboration paradigm;
ii) our method exhibits strong robustness under adverse conditions, surpassing state-of-the-art collaborative perception methods across all types of corruption. For instance, it shows a performance improvement of 4.9\% in cross sensor corruption at AP@0.5 and 4.02\% in crosstalk corruption at AP@0.7 on  OPV2V dataset.

\begin{figure}[h]
  \centering
   \includegraphics[width=1\linewidth]{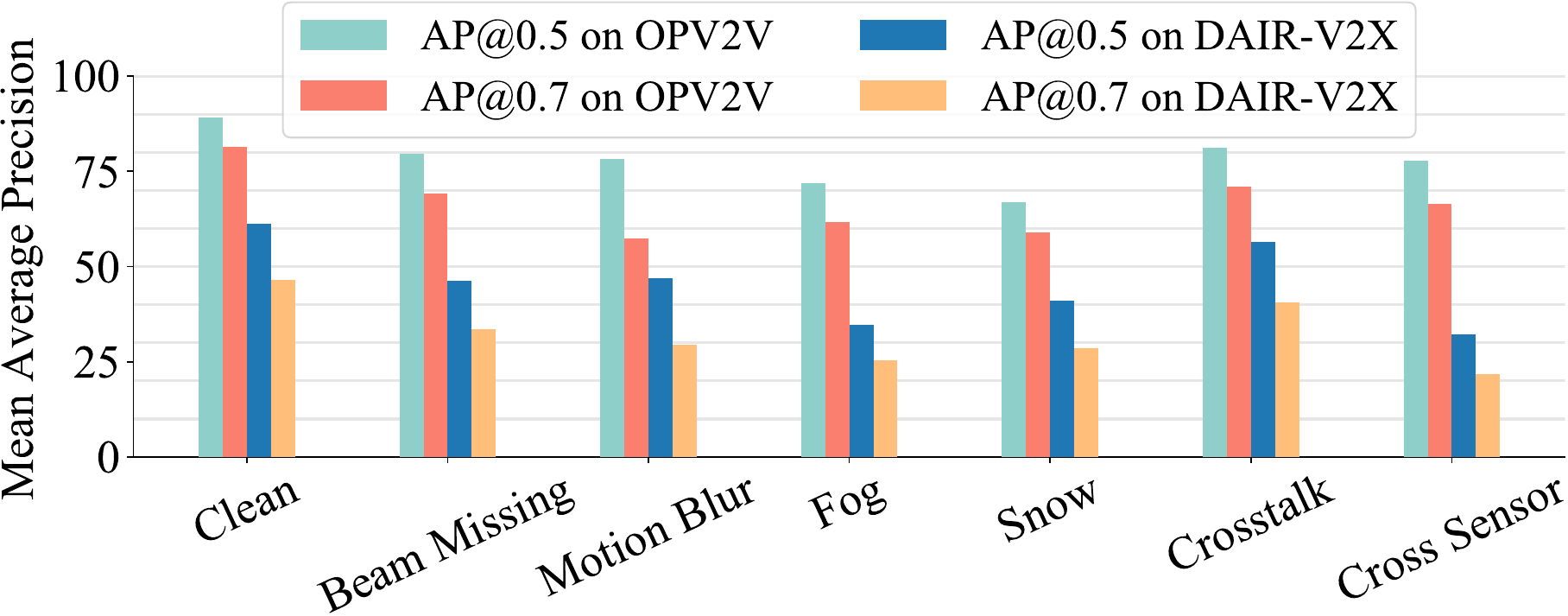}
   \caption{The average performance of all models under different corruption types.
   }
   \label{corruption}
   \vspace{-4mm}
\end{figure}

\textbf{Comparison of corruption types.} Table~\ref{tab:performance} and Figure~\ref{corruption} indicate  that all types of corruption lead to a decline in model performance, with weather-related corruption having  the most significant impact. For example, snow and fog conditions result in an average drop in AP@0.7 of over 20\% across all models, highlighting the threat of adverse weather to collaborative perception methods. Notably, under fog conditions, our model achieves  8.5\% higher AP@0.5 than the state-of-the-art on the OPV2V dataset, demonstrating its robustness. Additionally, motion blur poses a substantial challenge to all models, likely due to noise offsets exceeding the grid size. In contrast, most models show minor performance degradation under crosstalk corruption. This is mainly because the multi-agent environment of collaborative perception makes such corruption ubiquitous in the training dataset, increasing the models' resilience to it.

\begin{figure}[h]
  \centering
   \includegraphics[width=0.99\linewidth]{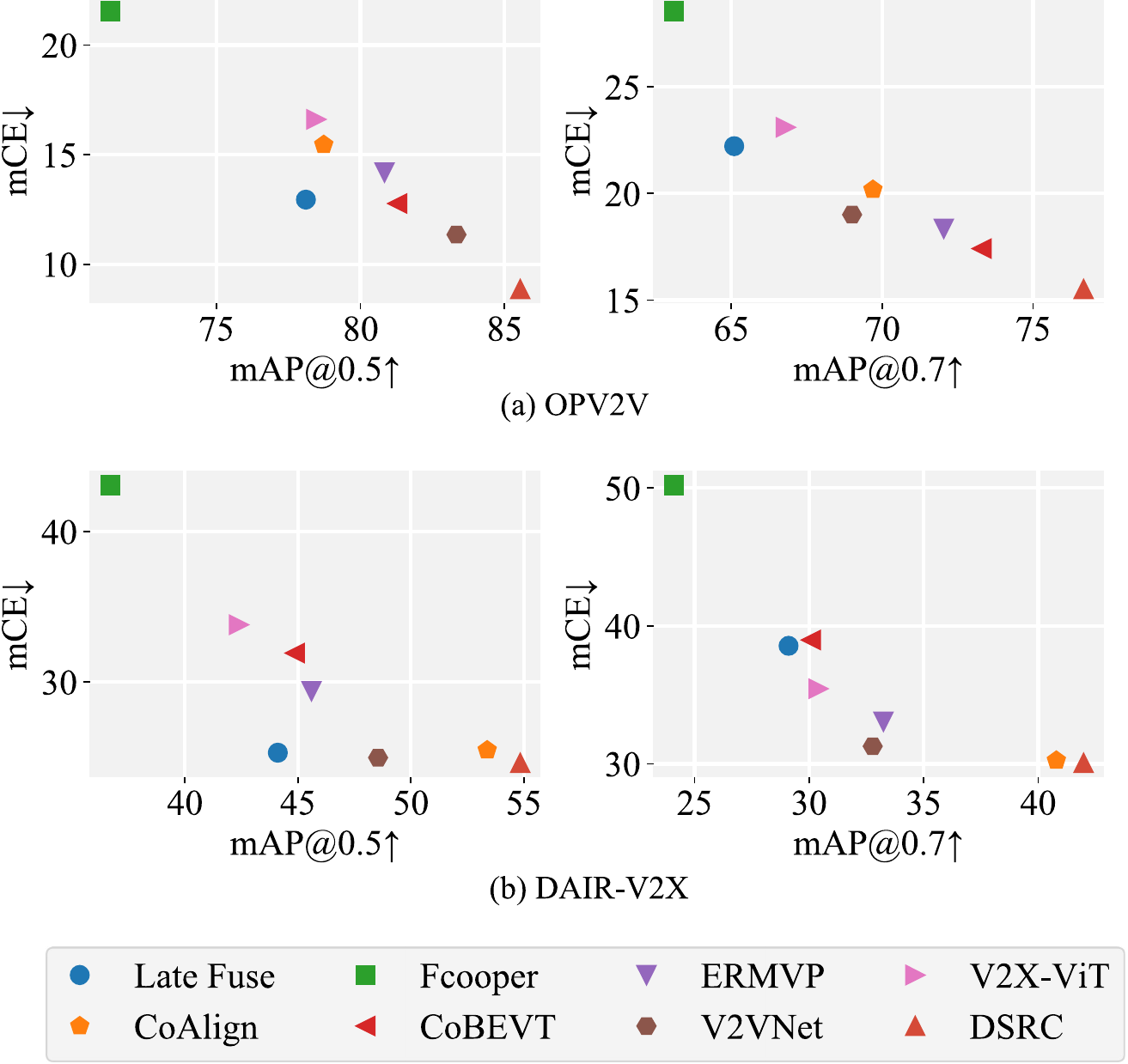}
   \caption{Benchmarking results of all models on the six robustness sets. The figure shows the mean corruption error (mCE) vs. the mean average precision (mAP).
   }
   \label{robustness}
   \vspace{-4mm}
\end{figure}

\textbf{Comparison of corruption robustness.}
Evaluation of model robustness under various corruption scenarios is crucial for achieving practical perception. To this end, we employ the mean corruption error (mCE) and the mean average precision (mAP) to assess the model's robustness. The mCE represents the average percentage of performance drop across different types of corruption, while the mAP indicates the average precision under these corruptions. As shown in Figure~\ref{robustness}, F-cooper exhibits the poorest model performance due to its simple maximum element selection method. In contrast, the Late Fusion  approach demonstrates better model robustness by avoiding the accumulation of errors from multi-agent perception features. Notably, DSRC achieves the highest accuracy and the least performance drop across both datasets, underscoring its superior robustness. The reasonable explanations are: (i) the proposed distillation framework effectively learns enhanced collaborative
representation in the latent space; (ii) the point cloud reconstruction module achieves a better fusion of crucial collaborative representation across agents.

\subsection{Ablation Studies}
\begin{table}[h] 
\setlength\tabcolsep{3.5mm}
\renewcommand\arraystretch{1.1}
\begin{small}
    \centering
    \begin{tabular}{ccc|cc}
    \hline
      SDD  &PCP   & REC   & OPV2V  & DAIR-V2X \\ \hline
         ~     & ~    & ~   & 79.75/70.54       & 53.11/40.12 \\ 
        \ding{51}    & ~    & ~   & 83.42/73.92          & 53.84/40.98 \\ 
       ~    & \ding{51} & ~   & 84.04/74.33  & 53.97/41.02 \\ 
      \ding{51}  & \ding{51} & ~  &84.83/75.99  & 54.28/41.37 \\
      \ding{51}  & \ding{51} & \ding{51} &\textbf{85.54/76.67} & \textbf{54.84/41.98} \\
      \hline
    \end{tabular}
\end{small}
      \caption{ Ablation study results of the proposed core designs on the both datasets. SDD: Sparse to Dense Distillation; PCP: Point Cloud Painting; REC: Point Cloud Reconstruction. The results are reported with average precision under all corruptions at IoU thresholds of 0.5 and 0.7.
      }
      \vspace{-2mm}
        \label{Tab3:Components}
\end{table}
\noindent
\textbf{Effect of Core Designs.} Table~\ref{Tab3:Components} details the contribution of each core design in our DSRC framework. The base model is a student model supervised solely by detection loss. We then assess the impact of each design by sequentially introducing: i) Sparse to Dense Distillation (SDD), ii) Point Cloud Painting (PCP), and iii) Point Cloud Reconstruction (REC). The consistent improvement in detection results across both datasets demonstrates the effectiveness of each introduced design. Notably, integrating all three components boosts detection performance by 5.79\% and 6.13\% on the OPV2V dataset for AP@0.5 and AP@0.7, respectively.
Due to space constraints, additional experiments are included in \textit{Appendix}.

\begin{figure}[h]
\subfigure[Point cloud]{
\begin{minipage}[b]{0.47\linewidth}
\centering
\includegraphics[width=0.99\linewidth]{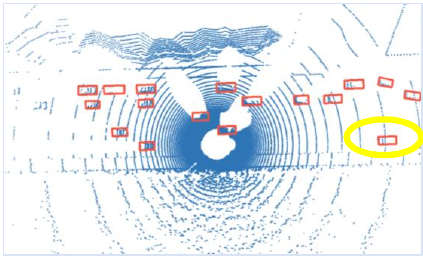}
\end{minipage}
}
\subfigure[Baseline]{
\begin{minipage}[b]{0.47\linewidth}
\centering
\includegraphics[width=0.99\linewidth]{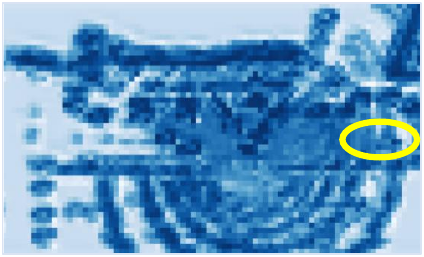}
\end{minipage}
}
\subfigure[Teacher]{
\begin{minipage}[b]{0.47\linewidth}
\centering
\includegraphics[width=0.99\linewidth]{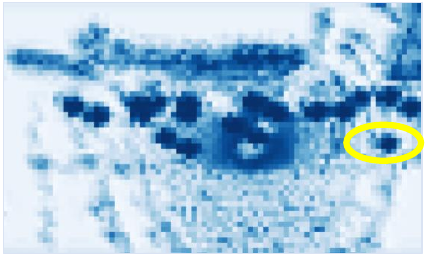}
\end{minipage}
}
\hspace{0.01mm}
\subfigure[Student]{
\begin{minipage}[b]{0.47\linewidth}
\centering
\includegraphics[width=0.99\linewidth]{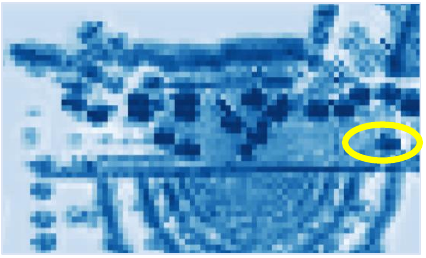}
\end{minipage}
}
     \vspace{-2mm}
	\caption{Visualisation of raw point cloud with features extracted by different models. \textcolor{red}{Red} 3D bounding boxes represent the ground truth. }
    \vspace{-4mm}
 \label{fig6}
\end{figure}

\subsection{Qualitative Evaluation}
The qualitative results are shown in Figure~\ref{fig6}. We visualize the original point cloud data alongside the corresponding feature maps of three types of models: baseline, teacher, and student. Here, we use a student model supervised solely by detection loss as the Baseline. The observations are as follows:
(i) the sparse-to-dense distillation framework successfully compensates for the sparse features of occluded or distant objects (circled in yellow), enabling the student model to produce denser and more robust features;
(ii) compared to the Baseline, the student model exhibits a more distinct separation between the background and object regions. This improvement is attributed to the semantic-guided approach that allows the student network to better extract meaningful semantic information and focus on perceptually critical features.
The visualization of object detection results can be found in the \textit{Appendix}.

\section{Conclusion}
This paper proposes DSRC, an innovative collaborative perception framework to enhance robustness against common corruptions. It considers a semantic-guided sparse-to-dense distillation framework to learn density-insensitive and semantic-aware collaborative representation effectively. Meanwhile, a feature-to-point cloud reconstruction approach is introduced to fuse critical perceptual information across agents better. Our method conducts supervised learning from multiple dimensions, including the original point cloud, latent features, and predictions, to stimulate more effective collaboration. Extensive experiments demonstrate that DSRC outperforms state-of-the-art collaborative perception methods in clean and corrupted scenarios. 
\section*{Acknowledgments}
This work is supported in part by 
the National Key Research and Development Program of China, Project No. 2024YFE0200700, Subject No. 2024YFE0200703,
the Specific Research Fund of the innovation Platform for Academicians of Hainan Province under Grant YSPTZX202314, 
the Shanghai Key Research Laboratory of NSAI, the Joint Laboratory on Networked Al Edge Computing,
Fudan University-Changan, and the National Natural Science Foundation of China (Grant No. 62250410368).
\bigskip
\bibliography{aaai25}

\newpage
\twocolumn

\maketitle
\section{ Additional Experiment Results and Details}
\subsection{Implementation details}
We implement all methods based on the OpenCOOD~\cite{opv2v} codebase. For the OPV2V dataset, the perception range is 281.6 m × 80 m, and for the DAIR-V2X dataset, the perception range is 201.6 m × 80 m. The size of the voxelization grid is 0.4 m × 0.4 m. We further downsample the feature map by a factor of 2 and reduce the feature dimension to 64 for efficient message sharing. The multi-scale feature attention fusion dimensions are \{64, 128, 256\}, and the Res2Net~\cite{gao2019res2net} layers consist of \{3, 5, 8\} blocks each. The confidence score threshold is set to 0.25, and non-maximum suppression (NMS) with an IoU threshold of 0.15 is applied to filter out overlapping detections.

\subsection{Additional Experiment Results}

\begin{table}[h] 
\setlength\tabcolsep{3.5mm}
\renewcommand{\arraystretch}{1.1}
\begin{small}
\centering
\begin{tabular}{c|c|c}
\hline
Model                   &  \multicolumn{1}{c|}{Maximum (vanilla)}  &  \multicolumn{1}{c}{Maximum + DSRC }        \\ \hline
\multicolumn{3}{c}{OPV2V}      \\ \hline
Clean                &87.40/79.40     &91.51/86.07            \\
Beam Missing                 & 75.65/65.91     &83.16/74.97           \\
Motion Blur               &73.44/48.30    &80.26/58.45     \\
Fog                  &63.76/53.52   &71.48/65.38                   \\
Snow              & 59.60/53.21   &59.72/54.91                        \\
Crosstalk                & 63.76/53.52     &76.45/69.48               \\
Cross Sensor                 &75.65/65.91     &81.47/71.86          \\ \hline
\multicolumn{3}{c}{DAIR-V2X}      \\ \hline
Clean                &58.23/41.74     &59.50/42.53              \\
Beam Missing                 &40.11/27.76     &41.16/28.50             \\
Motion Blur               & 33.71/17.36   &35.41/18.26                \\
Fog                  & 27.03/18.64  & 27.56/19.36                  \\
Snow              & 20.88/11.69 &22.56/12.63                       \\
Crosstalk                &50.78/33.97   &52.15/34.94               \\
Cross Sensor                 &26.29/17.48      &27.09/18.08             \\

\hline  
\end{tabular}
\end{small}
\caption{Performance comparison using position-wise maximum fusion strategy between the vanilla model and the model enhanced with our framework. The results are reported in AP@0.5/0.7.}
\vspace{-4mm}
\label{fusion_strategy}
\end{table}

\begin{table}[h] 
\setlength\tabcolsep{2.5mm}
\renewcommand{\arraystretch}{1.1}
\begin{small}
\centering
\begin{tabular}{c|c|c}
\hline
Model                   &  \multicolumn{1}{c|}{No Collab.(Vanilla)}  &  \multicolumn{1}{c}{No Collab + DSRC}        \\ \hline
\multicolumn{3}{c}{OPV2V}      \\ \hline
Clean                     &78.85/65.05      &85.24/71.73            \\
Beam Missing              &63.50/48.01      &68.31/54.68            \\
Motion Blur               &61.99/36.79      &71.66/49.49            \\
Fog                       &59.88/49.38      &60.77/52.98            \\
Snow                      &57.56/45.48      &64.41/52.45            \\
Crosstalk                 &73.73/58.47      &79.94/65.66            \\
Cross Sensor              &60.64/44.00      &68.51/53.08            \\   \hline
\multicolumn{3}{c}{DAIR-V2X}      \\ \hline
Clean                     &54.19/42.43      &55.31/46.33            \\
Beam Missing              &43.11/33.62      &44.37/37.06            \\
Motion Blur               &38.74/24.53      &47.22/36.03            \\
Fog                       &26.19/19.75      &36.32/30.66            \\
Snow                      &35.75/26.82      &41.22/33.95            \\
Crosstalk                 &47.83/35.05      &52.47/43.65            \\
Cross Sensor              &30.01/22.70      &30.60/24.90            \\

\hline  
\end{tabular}
\end{small}
\caption{Comparison of perception performance of single-vehicle modes in the no collaboration state. The results are reported in AP@0.5/0.7.}
\vspace{-2mm}
\label{single_vehicle_modes}
\end{table}

\begin{figure}[h] 
\centering
\includegraphics[scale=0.20]{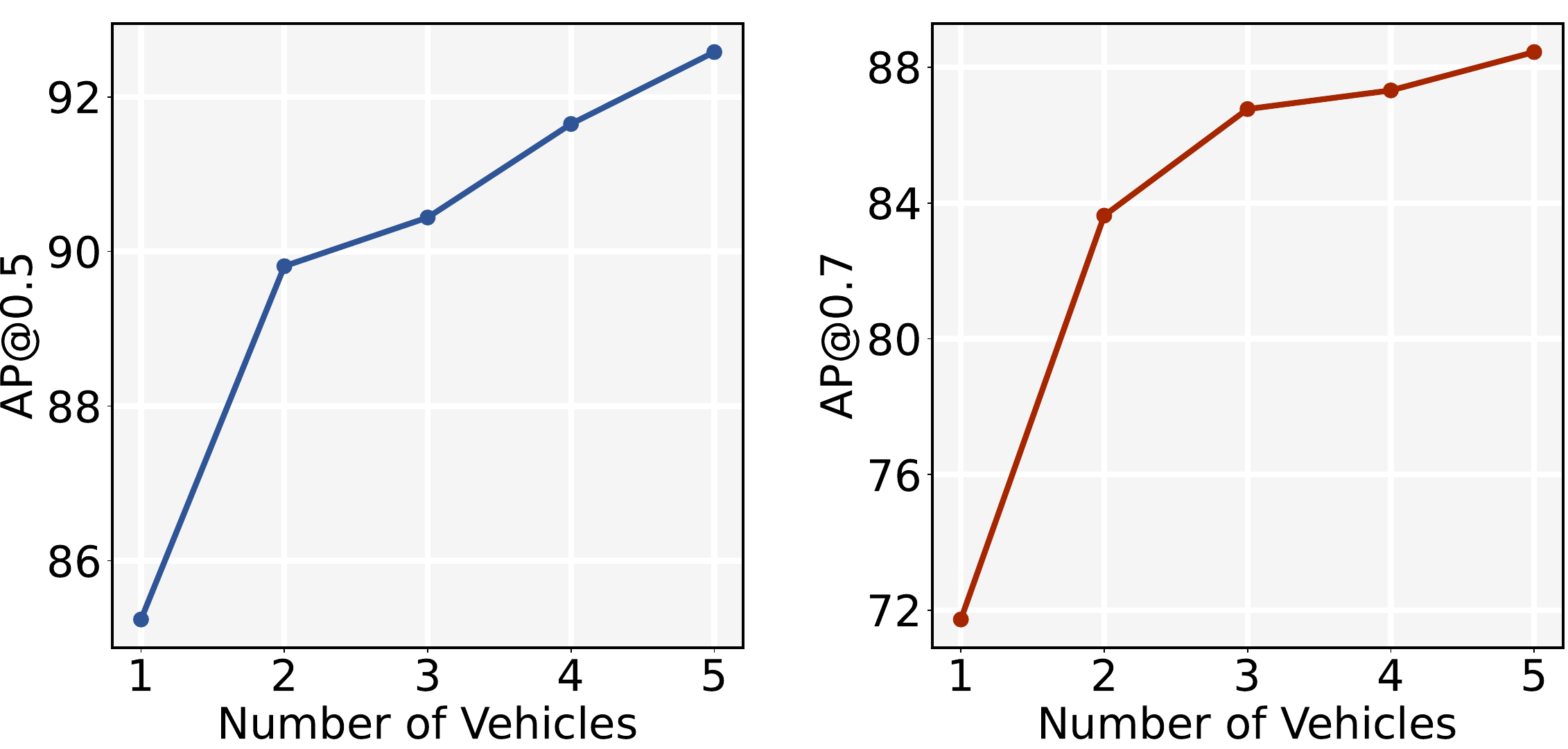} 
\caption{Ablation study on number of agents.
} 
\label{number} 
\end{figure}

\textbf{Number of agents.}
In this part, we evaluate the impact of the number of collaborative agents on the performance of DSRC. As depicted in Figure~\ref{number}, it is observed that an increase in the number of collaborators typically leads to an improvement in performance. However, this improvement becomes marginal when the number of agents exceeds three.

\textbf{Ablation studies with fusion strategy.} To evaluate the effectiveness of our framework, we experimented with different feature fusion strategies. Specifically, we replaced the multi-scale feature attention fusion strategy with position-wise maximum fusion.  Table~\ref{fusion_strategy} presents the performance comparison of the models (vanilla and ours) using position-wise maximum fusion on the OPV2V and DAIR-V2X datasets.
Our enhanced model using the distillation framework consistently outperforms the vanilla model across different weather conditions in both datasets. For instance, in clean weather conditions, the AP@0.7 improved by 6.67\% on the OPV2V dataset and by 0.79\% on the DAIR-V2X dataset. Similarly, under foggy conditions, the AP@0.7 increased by 11.86\% on the OPV2V dataset and by 0.72\% on the DAIR-V2X dataset. These improvements demonstrate the robustness and flexibility of our method, which effectively adapts to various feature fusion strategies.

\textbf{Comparison of no collaborative state.}
Table~\ref{single_vehicle_modes} compares the average precisions of the models (vanilla and ours) in a no collaborative state at IoU thresholds of 0.5 and 0.7 for various weather conditions on the OPV2V and DAIR-V2X datasets. The results indicate that the enhanced No Collaboration model, leveraging our distillation framework, significantly outperforms the vanilla model across different weather conditions in both datasets.
For example, in clean weather conditions, the AP@0.7 improved by 6.68\% on the OPV2V dataset and by 3.90\% on the DAIR-V2X dataset. This demonstrates that our distillation framework enhances the feature representation capability of the single-vehicle perception model. Furthermore, under foggy conditions, the AP@0.7 increased by 3.60\% on the OPV2V dataset and by 10.91\% on the DAIR-V2X dataset, indicating that our approach also significantly enhances the robustness of the single-vehicle perception model under adverse weather conditions.
These results confirm that our distillation framework not only improves performance in collaborative settings but also significantly boosts the robustness and accuracy of single-vehicle perception in various  conditions.

\section{Comprehensive Description of phases}
\subsection{Metadata Sharing and Feature Extraction}
\textbf{Metadata Sharing.}
Following the settings of existing works~\cite{opv2v,wang2023core,v2vnet},  we define a agent as the ego agent and designate other agents within a 70 m communication range~\cite{xu2022v2x} as collaborators. This approach aims to augment the perception capabilities of the ego agent by integrating complementary information from both its local observations and the features shared by collaborators. In practice, the ego agent broadcasts its metadata (e.g., poses, extrinsics, and sensor type) via the established Vehicle-to-Everything communication channels. The transmission delay for this metadata is negligible, given the minimal size and volume of the communication. Upon receiving the pose of the ego agent, all the other connected agents nearby will project their own LiDAR point clouds to the ego-agent’s coordinate system before feature extraction.

\textbf{Feature Extraction}. 
This function is to extract informative features from the point clouds and convert them into Bird’s Eye View (BEV). We assume that there are $N$ agents perceiving the environment, $X_i$ is the 3D point clouds observed by the $i$-th agent, the features of the $i$-th agent is the features of the $i$-th agent is obtained as
$
\mathbf{F}_i = f_{\text {encode}}(\mathbf{X}_i) \in \mathbb{R}^{H \times W \times C},
$
where $f_{\text {encode }}$ is an encoder to extract features and $H$, $W$, $C$ are the features' height, weight and channel.
We adopt the anchor-based PointPillar method as introduced in ~\cite{pointpillars} for extracting informative features. This method is preferred for real-world deployment over other 3D detection backbones (e.g., SECOND~\cite{yan2018second}, PIXOR~\cite{yang2018pixor}, VoxelNet~\cite{zhou2018voxelnet}) due to its lower inference latency and efficient memory usage.
Specifically, the point cloud in 3D space is discretized into a uniformly distributed grid in the 2D plane, each grid is a stacked pillar tensor, then scattered to a 2D pseudo-image which is also called feature. Then the extracted feature is output to the filter and merge feature sampling module.

\subsection{Multi-scale Feature Fusion}
The purpose of this module is to enhance the visual representation of the ego-agent by fusing features from nearby agents. To better capture complementary spatial information, we use a multis-cale feature attention fusion method to purposefully fuse perceptual information from different agents, thus producing information-rich and robust features. Specifically, for the features $\mathbf{F}_{j \rightarrow i}^{(\ell)} $ of the $j$th agent at the $\ell$th scale, the fusion process can be expressed as
\begin{equation}
\begin{aligned}
& \mathbf{F}_{j \rightarrow i}^{(\ell+1)}  = d_{\ell}\left(\mathbf{F}_{j \rightarrow i}^{(\ell)}\right), \ell=1,2, \cdots, L \\
& \mathbf{H}_{i}^{\ell}=\sum_{j \in \mathcal{N}_i \cup\{i\}}\operatorname{softmax}\left(\frac{ \mathbf{F}_i^{(\ell)} \mathbf{F}_{j \rightarrow i}^{(\ell)\top} }{\sqrt{d_k}}\right) \mathbf{F}_{j \rightarrow i}^{(\ell)} \\
&\mathbf{\widetilde H}_i  =\|\left[\mathbf{H}_i^{(1)}, u_2\left(\mathbf{H}_i^{(2)}\right), \cdots, u_L\left(\mathbf{H}_i^{(L)}\right)\right],
\end{aligned}
\end{equation}
where $d_{\ell}$ is the $\ell$th convolution layer with downsampling by 2, and $\mathcal{N}_i$ is the neighbours of the $i$th agent. $\mathbf{F}_{i}^{(\ell)} $ is the feature of ego-agent at the $\ell$th scale. $u_{\ell}(\cdot)$ is a transposed convolution upsampling operator for the $l$th scale and $\|$ is a concatenation operation along the feature channel dimension. $\mathbf{\widetilde H}_i$ is the fused featureS. Finally, the fused features are fed into a decoder to obtain the perception result.

\begin{figure*}[t] 
	\centering
	\subfigure[Fcooper]{
		\begin{minipage}[t]{0.24\linewidth}
			\centering
			\fbox{\includegraphics[width=1.55in]{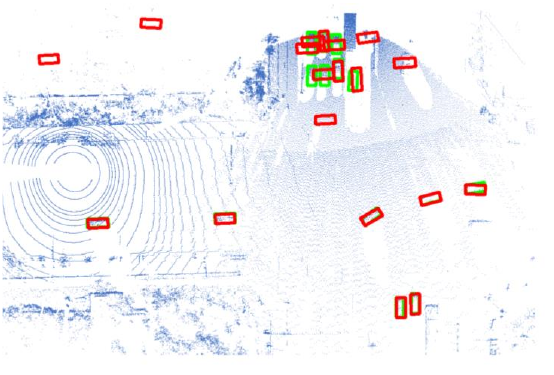}}\\
			\vspace{0.15cm}
			\fbox{\includegraphics[width=1.55in]{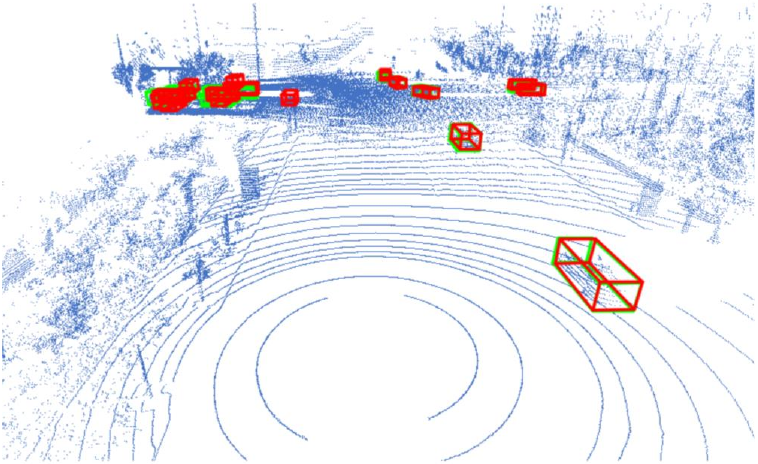}}\\
			\vspace{0.15cm}
   			\fbox{\includegraphics[width=1.55in]{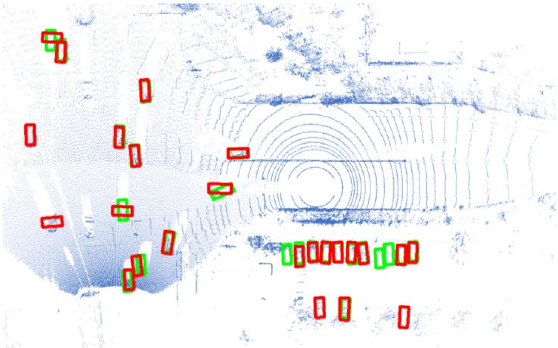}}\\
			\vspace{0.15cm}
   			\fbox{\includegraphics[width=1.55in]{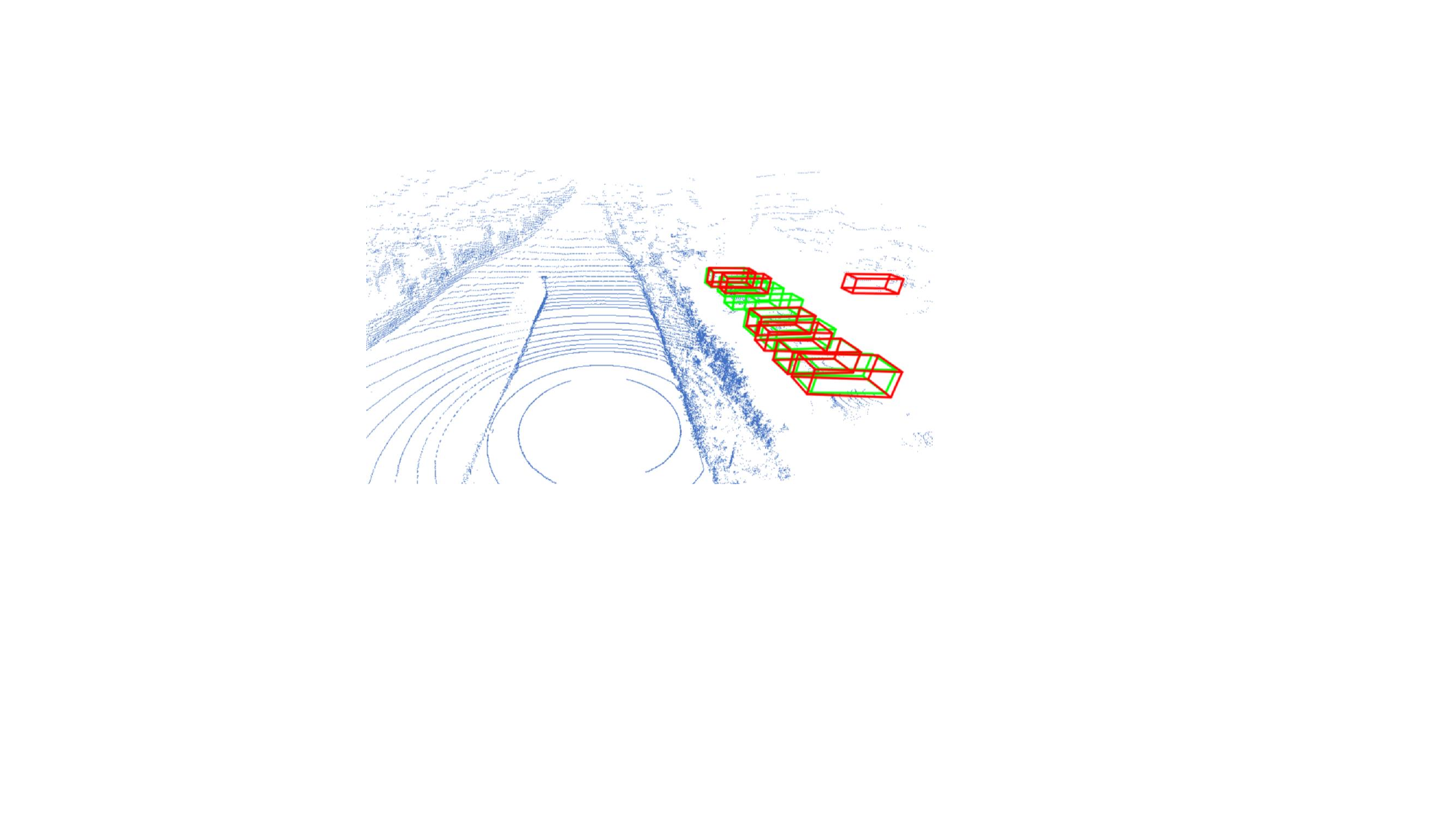}}\\
			\vspace{0.15cm}
		\end{minipage}%
	}%
	\subfigure[V2X-ViT]{
		\begin{minipage}[t]{0.24\linewidth}
			\centering
			\fbox{\includegraphics[width=1.55in]{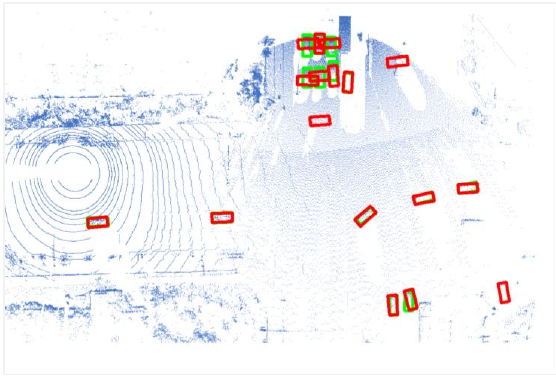}}\\
			\vspace{0.15cm}
			\fbox{\includegraphics[width=1.55in]{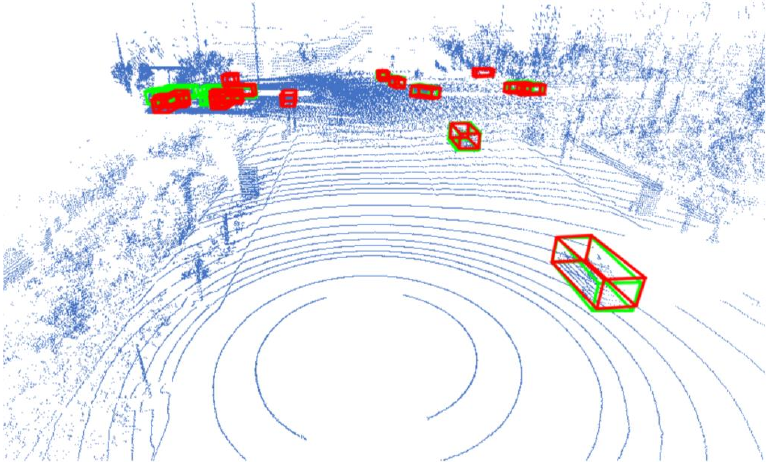}}\\
			\vspace{0.15cm}
   			\fbox{\includegraphics[width=1.55in]{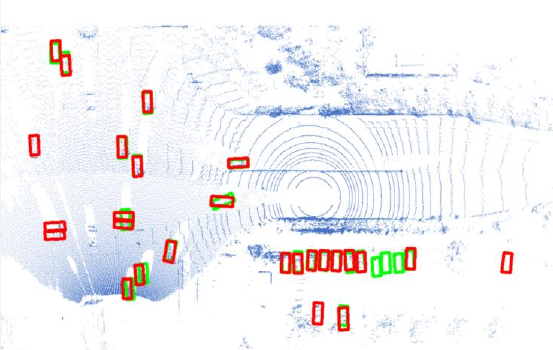}}\\
			\vspace{0.15cm}
   			\fbox{\includegraphics[width=1.55in]{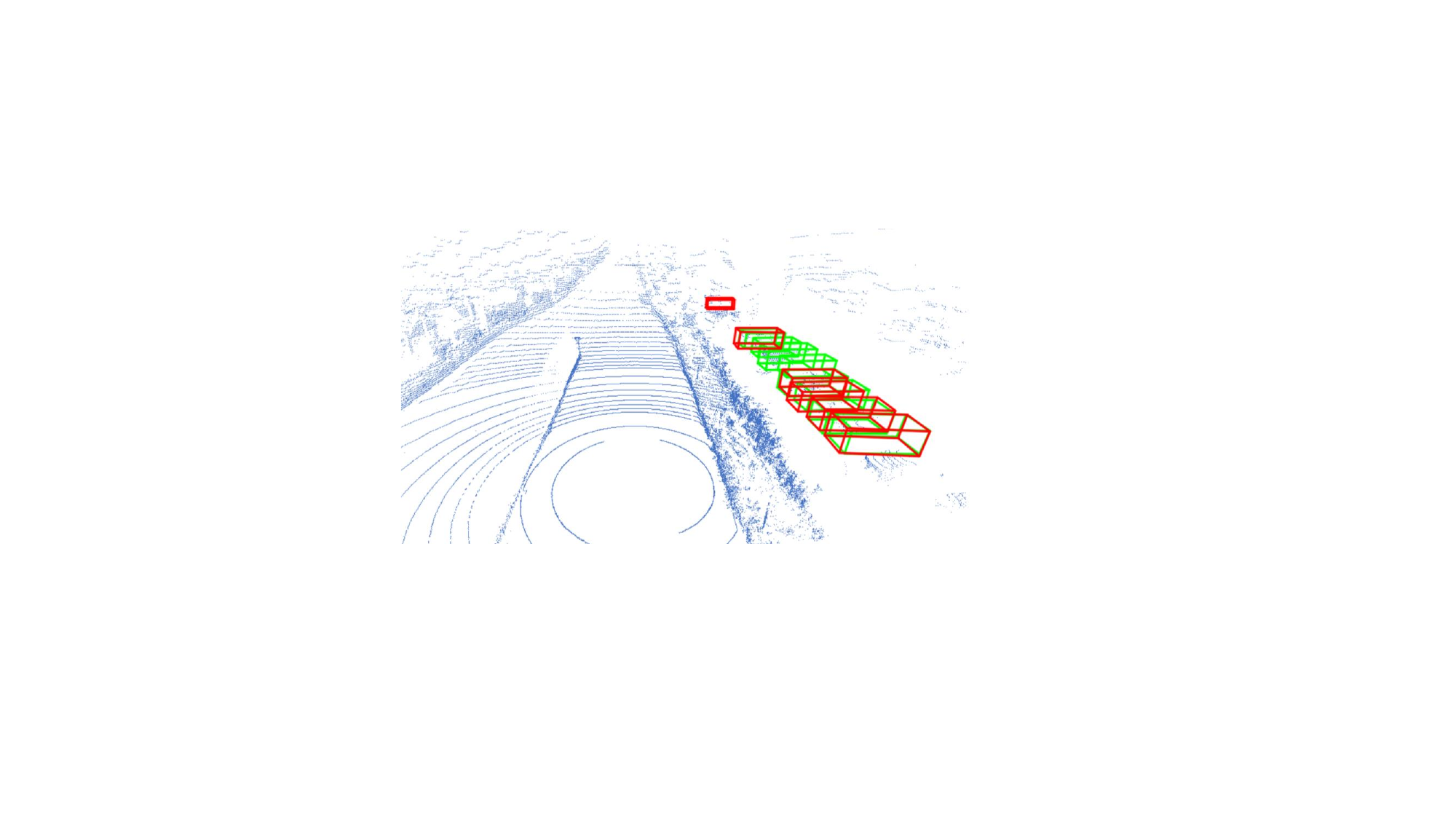}}\\
			\vspace{0.15cm}
		\end{minipage}%
	}%
	\subfigure[CoBEVT]{
		\begin{minipage}[t]{0.24\linewidth}
			\centering
			\fbox{\includegraphics[width=1.55in]{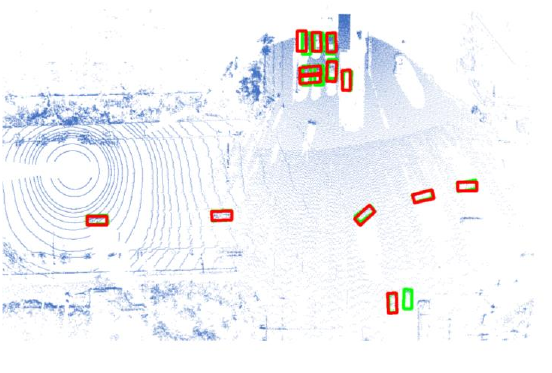}}\\
			\vspace{0.15cm}
			\fbox{\includegraphics[width=1.55in]{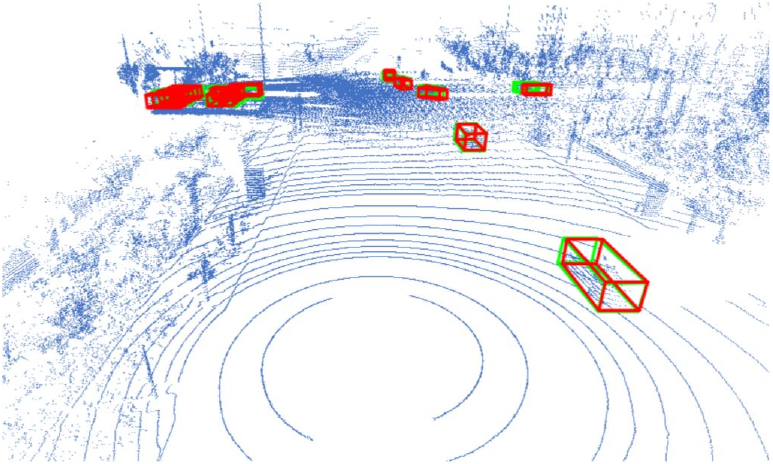}}\\
			\vspace{0.15cm}
   			\fbox{\includegraphics[width=1.55in]{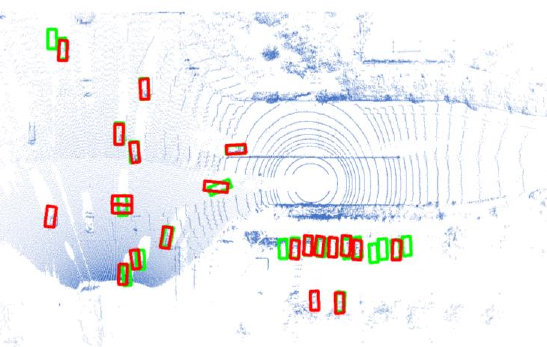}}\\
			\vspace{0.15cm}
   			\fbox{\includegraphics[width=1.55in]{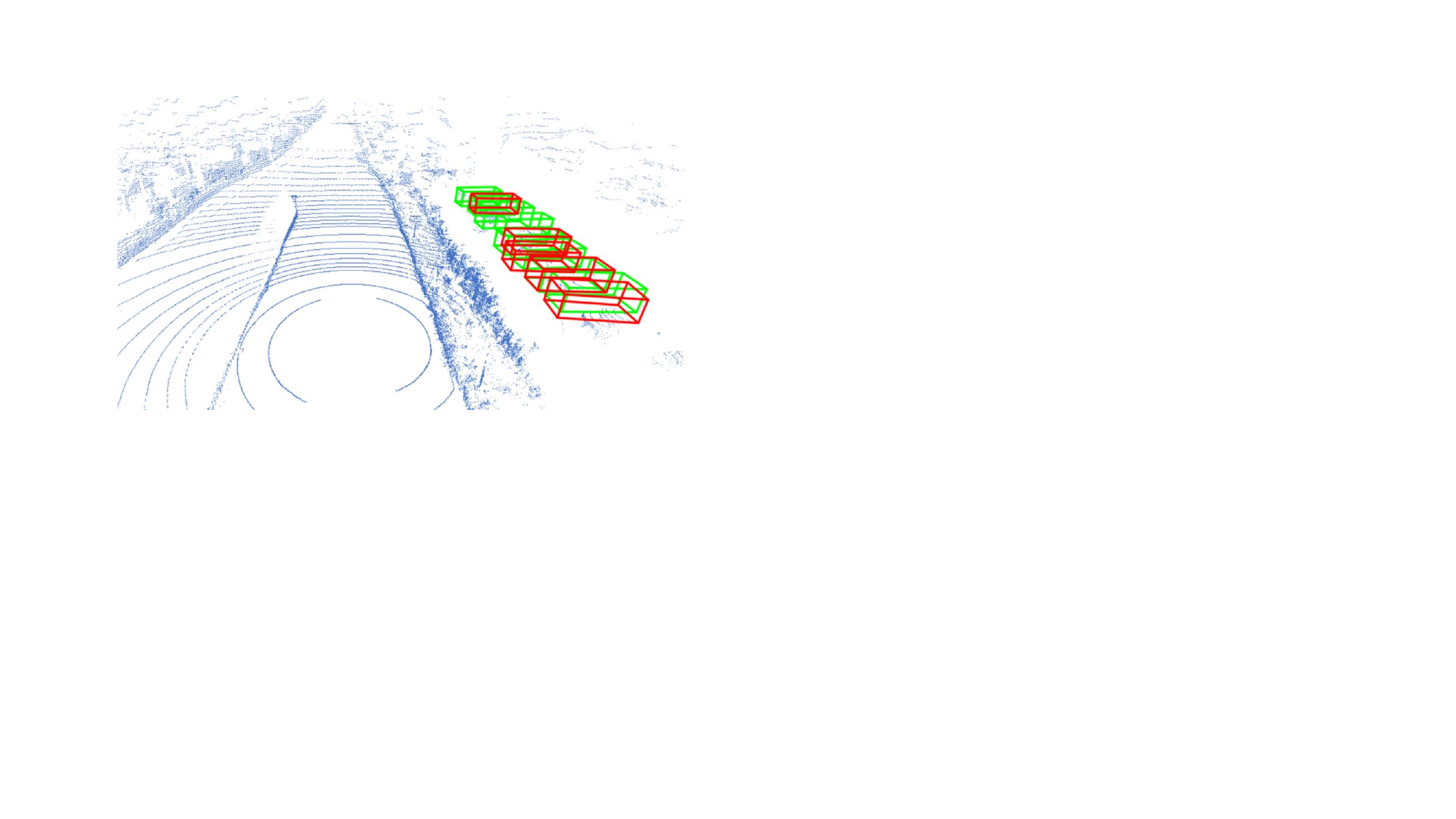}}\\
			\vspace{0.15cm}
		\end{minipage}%
	}%
	\subfigure[DSRC]{
		\begin{minipage}[t]{0.24\linewidth}
			\centering
			\fbox{\includegraphics[width=1.55in]{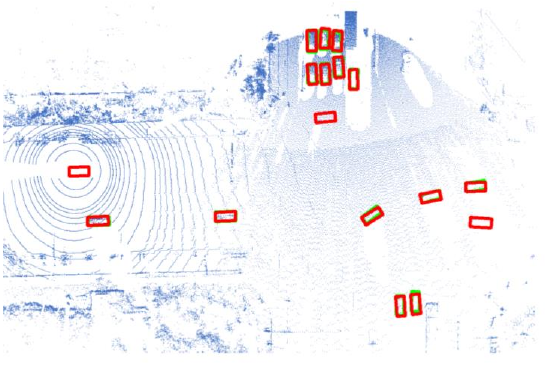}}\\
			\vspace{0.15cm}
			\fbox{\includegraphics[width=1.55in]{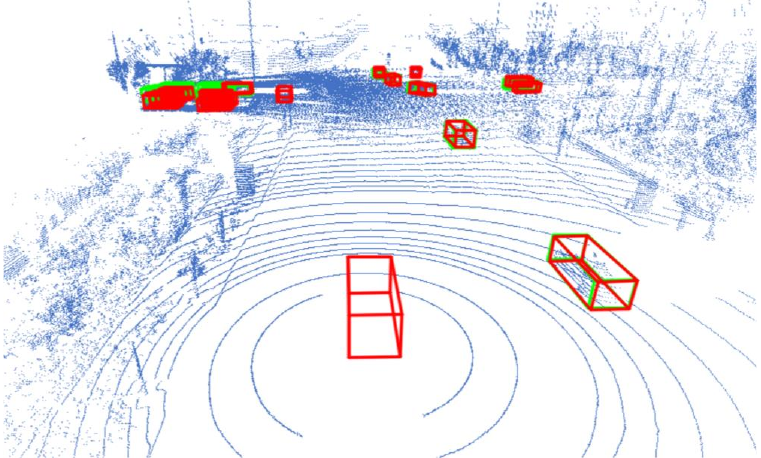}}\\
			\vspace{0.15cm}
   			\fbox{\includegraphics[width=1.55in]{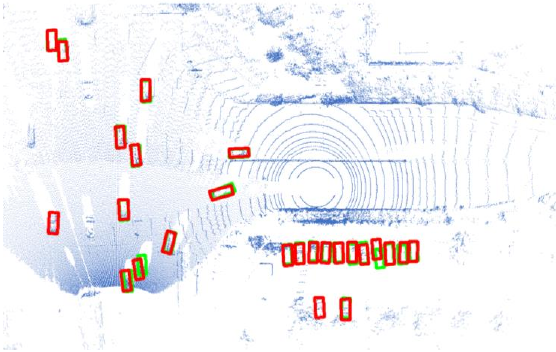}}\\
			\vspace{0.15cm}
   			\fbox{\includegraphics[width=1.55in]{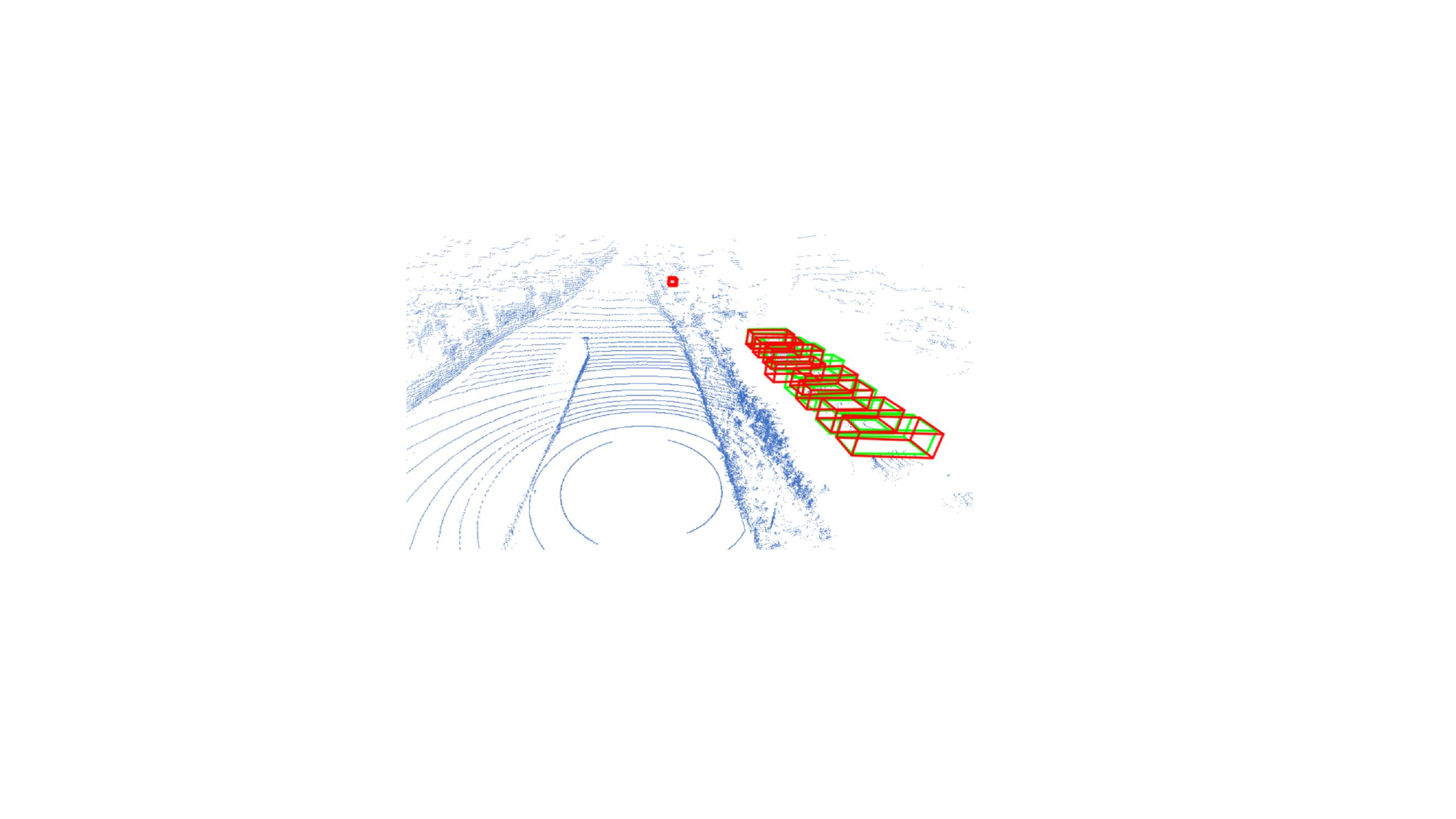}}\\
			\vspace{0.15cm}
		\end{minipage}%
	}%
	\caption{Visualization of collaborative perception results on a roadway in 3D view and BEV view, with \textcolor{green}{green} and \textcolor{red}{red} 3D bounding boxes representing the ground truth and prediction, respectively. DSRC produces more accurate detection outputs.}
	\label{fig4}
 \vspace{-2mm}
\end{figure*}

\section{More Details of Common Corruptions}
In this section, we provide technical details about the six common corruptions in the paper.

\textbf{Beam Missing.} 
The phenomenon of LiDAR beam missing arises when certain laser beams fail to propagate to the target or are not received due to environmental interference, surface properties of the target, or limitations in sensor performance, resulting in void regions within the point cloud. To emulate this effect, we randomly selected 16 beams from the original point cloud and excluded all points associated with these beams.

\textbf{Motion Blur.} 
When LiDAR sensors capture data during rapid motion, the movement of the sensor can cause fixed object position information in the point cloud to shift or blur. Additionally, when the target object moves during LiDAR scanning, the point cloud data captured by LiDAR will be affected by the object's trajectory, resulting in stretching or deformation of the object's geometry in the point cloud. To simulate motion blur, we added jitter noise with a standard deviation of 0.2 to each point coordinate.

\textbf{Fog.} Water droplets and tiny particles in fog cause scattering and absorption of laser beams, leading to attenuation of laser signal energy during transmission and affecting the signal-to-noise ratio. Additionally, multiple scattering effects under foggy conditions cause LiDAR to receive return signals from multiple paths, increasing data processing complexity and potentially introducing additional measurement errors. We employed a physically effective fog simulation method~\cite{hahner2021fog} create fog-damaged data. Specifically, we decomposed the spatial impulse response in fog into hard target and soft target terms. The hard target term represents an attenuated version of the original clear-weather response, while the soft target term represents echoes reflected by fog particles. Finally, we updated the point cloud using the maximum criterion and a noise factor.

\textbf{Snow.}  Following~\cite{hahner2022lidar}, we adopted a linear system model to simulate the transmission and reception power of LiDAR pulses. By explicitly sampling snow particles as opaque spheres, we simulated various intensities of snowfall conditions. During sampling, we followed an exclusion principle to ensure particles did not intersect. For each LiDAR beam, we calculated the set of intersecting particles, considered potential occlusion factors, derived the cross-sectional angle of beams reflected by each particle, and computed the received power under snowy conditions.

\textbf{Crosstalk.} Point cloud crosstalk refers to measurement errors and data inaccuracies caused by signal interference in multi-sensor systems during point cloud generation. This interference is typically due to multiple reflections, overlapping sensor fields of view, and other factors, leading to the appearance of false targets and data noise. To simulate this, we randomly sampled a subset of points proportional to 0.01 from the original point cloud and added 3m Gaussian noise.

\textbf{Cross Sensor.} Effectively accommodating the heterogeneity of LiDAR data across different devices is crucial to ensuring satisfactory performance of models on various device configurations. However, previous studies often directly applied settings from different datasets for such comparisons, where the domain characteristics of these datasets further limited direct robustness comparisons. We followed previous methods~\cite{wei2022lidar} by first deleting points from specific beams in the point cloud and sub-sampling from each beam to generate cross sensor data settings.

\section{The Visualization of Object Detection Results}
To intuitively validate the effectiveness of our model, we present the detection results of Fcooper~\cite{F-cooper}, V2X-ViT~\cite{xu2022v2x}, CoBEVT~\cite{CoBEVT}, and DSRC under the clean setting of the real-world dataset DAIR-V2X, as shown in Figure~\ref{fig4}. From Figure~\ref{fig4}, it is evident that by comparison with the other three models, our model can achieve a more comprehensive and accurate detection with a smaller shift from the ground truth. Meanwhile, it is observed that the first three models are prone to miss detection and erroneous detection in the scene edges or crowded areas. In contrast, our model can still generate comprehensive and accurate bounding boxes in the above cases at the proper target location. The above results illustrate that our model can learn more comprehensive and semantically rich information in the scene, which further validates the effectiveness of our proposed distillation framework.

\end{document}